\documentclass[11pt]{article} % For LaTeX2e
\usepackage{rldmsubmit,palatino}
\usepackage{graphicx}
\usepackage{hyperref}
\usepackage{algorithm}
\usepackage{algpseudocode}
\usepackage{wrapfig}
\usepackage{subfig}
\usepackage{xcolor}
\usepackage{natbib}

\title{\textbf{Does DQN \textit{really} learn?\\ Exploring adversarial training schemes in Pong}}

\author{
Bowen He\\
Department of Computer Science\\
Brown University\\
Providence, RI 02912\\
\texttt{bowen\_he@brown.edu}
\And
Sreehari Rammohan\\
Department of Computer Science\\
Brown University\\
Providence, RI 02912\\
\texttt{sreehari@brown.edu}
\And
Jessica Forde\\
Department of Computer Science\\
Brown University\\
Providence, RI 02912\\
\texttt{jessica\_forde@brown.edu}
\And
Michael L.\ Littman\\
Department of Computer Science\\
Brown University\\
Providence, RI 02912\\
\texttt{mlittman@cs.brown.edu}
}
\begin{document}

\maketitle
\begin{abstract}

  In this work, we study two self-play training schemes, \emph{Chainer} and \emph{Pool}, and show they lead to improved agent performance in Atari Pong compared to a standard DQN agent---trained against the built-in Atari opponent. To measure agent performance, we define a robustness metric that captures how difficult it is to learn a strategy that beats the agent's learned policy.Through playing past versions of themselves, Chainer and Pool are able to target weaknesses in their policies and improve their resistance to attack. Agents trained using these methods score well on our robustness metric and can easily defeat the standard DQN agent. We conclude by using linear probing to illuminate what internal structures the different agents develop to play the game. We show that training agents with Chainer or Pool leads to richer network activations with greater predictive power to estimate critical game-state features compared to the standard DQN agent. 
 
\end{abstract}


\startmain 

\section{Introduction}

Since the introduction in 2013, DQN~\citep{DQN} 
has been accepted as outperforming human-level play in a range of Atari games. However, a fundamental question about what these agents truly learn remains: Are they picking up on gameplay techniques and fundamental scoring concepts, or are they merely exploiting idiosyncrasies in the game dynamics? As an example, a small change to the opponent policy can totally confuse a standard DQN-trained agent because it overfits to the opponent gameplay weaknesses instead of learning the essential skills of the game~\citep{witty21}.  Furthermore, are there ways of training Deep RL agents that can facilitate their learning of the essential rules of the game, leading to more robust policies? We aim to answer these questions in the game of Pong, showing that adversarial training schemes can extract more context-relevant information leading to stronger policies. 

\section{Background}

The environment is modeled as a a time-homogeneous Markov Decision Process $(X , A, R, P, \gamma)$. As usual, $X$ and $A$ represent state and action spaces, respectively, while the transition kernel $P(\cdot|x,a)$ defines the dynamics of the environment. Reward function $R(x,a)$ provides a signal to direct learning. Deep Q Networks, or DQN, is an approach to learning to optimize behavior in such environments. It applies several strategies to stabilize learning compared to naive learning approaches: replay buffer sampling, a target network, and frame pre-processing tricks. 

Recent years have seen an uptick in the attention researchers have paid to questions of generalization of reinforcement-learning models. \citet{AssessingGeneralization} found that the vanilla deep RL algorithms have better generalization performance than specialized schemes that were proposed specifically to tackle generalization. In addition, \citet{SchemaNetworks} argued that generalizing from limited data and learning causal relationships are essential abilities on the path toward general intelligent systems. 

Probe techniques used by \citet{BERT} were useful to determine where different kinds of linguistic information is encoded in a large language model by correlating network activations with known properties of inputs. We apply these methods to our work, using linear probing to investigate the ability of the learned networks to support a heuristic task that well-trained agents should be able to perform well on.

Up until now, DQN in Pong has been trained non-adversarially, where a single agent competes against a static opponent modelled by the computer. The thesis of this paper is that adversarially trained agents are better than those trained non-adversarially in that they should develop policies prepared to handle a broader array of scenarios.

We observe in the training of a \emph{standard DQN agent} (used to refer a learner trained non-adversarial) in Pong that the agent learns to exploit weaknesses in the built-in Atari opponent---after a few million training steps, it perfects a ``kill shot'' aimed always at the upper or lower quadrants of the board that the built-in Atari agent cannot return. In our work, we change the dynamics of the opponent, and we demonstrate that doing so pushes the learner to focus on learning a more generalizable understanding of the game itself, robustly improving its skills in the process. 

\section{Methodology}

Our adversarial training schemes pit the current agent against a previous version of itself. The key difference between the two algorithms we propose is \textit{which} previous agent the current agent faces. We used the Gym Retro Atari implementation of Pong, which allows us to control both the left and right paddles independently. We chose Pong as a test bed because it is challenging, but small scale, allowing many training runs in a short amount of time.

In addition, Pong is well suited to adversarial training because the board is vertically symmetric, allowing us to easily create self-playing agents using a single policy. Specifically, if an agent is trained on one side, we can symmetrically reflect the game board as input to the agent to apply the same policy to the opposite side. To avoid overfitting the agent to a single side of the board, we randomly assign the agent a game side to play periodically.
% every fixed number of steps. 
Under this uniform setting, we empirically tested two methods of self-play, \emph{Pool} and \emph{Chainer}, described in the following sections. We use the standard DQN agent as a baseline and our empirical results show that it is much weaker than the pool and chainer agents.

\subsection{Pool}
   
Pool, described in Algorithm~\ref{alg:pool}, plays 
% the current iteration of the agent 
against a randomly selected previous agent from a queue of fixed size. % Every fixed number of steps, 
% MLL zzz: we can say how many, but just saying "fixed" doesn't seem very informative.
Periodically,
the current version of the agent is frozen and appended to the queue, creating a set of most up-to-date adversaries to play against. 
% Similarly, 
The opponent is updated periodically by sampling from this queue. In practice, we set the queue size to $5$ and add new agents to the pool every $250,000$ steps.

\subsection{Chainer}

Chainer, described in Algorithm~\ref{alg:chainer}, plays the current version of the agent, $A^{n}$, solely against its immediate predecessor, $A^{n-1}$. When $A_n$ achieves a fixed evaluation threshold, defined here as getting above an average score of $15$ over $10$ matches, the current agent $A^{n}$ replaces the opponent's policy and a successor $A^{n+1}$ will continue to be trained from the policy of $A^{n}$. We initialize the first member of the chain, $A^{0}$, with a standard DQN agent trained for $45$ million steps. Agent $A^{1}$ is trained from scratch (\textit{tabla rasa}) to play against agent $A^{0}$. This training approach allows the agent to continually improve its skill without overfitting to a specific opponent (which would stagnate the skill learning of the agent). 
% This algorithm can be thought of as a modified version of the Hoffman-Karp. \\

\begin{minipage}{0.45\textwidth}
\begin{algorithm}[H]
\caption{Pool Self-Play Algorithm}
\label{alg:pool}
\begin{algorithmic}
\State $P \gets p$, pool update frequency
\State $S \gets s$, step counter
\State Initialize \emph{Agent}
\State Initialize \emph{Env}
\State Initialize \emph{Pool} of fixed size
\State Initialize \emph{Opponent}
\While{True} 
    \State  \emph{interact}(\emph{Agent}, \emph{Opponent}, \emph{Env})
    \State \emph{update}(\emph{Agent})
    \If {$ S\%P == 0$}
        \State \emph{Pool} $\gets$ \emph{freeze}(\emph{Agent})
        \State \emph{Opponent} $\gets$ \emph{sample}(\emph{Pool})
    \EndIf
    \State $ S += 1$
\EndWhile
\end{algorithmic}

\end{algorithm}
\end{minipage}
\hfill
\begin{minipage}{0.45\textwidth}
\begin{algorithm}[H]
\caption{Chainer Self-Play Algorithm}\label{alg:chainer}
\begin{algorithmic}
\State $S \gets 0$, the number of steps
\State $C \gets c$, the evaluation frequency
\State $i \gets 1$, current length of chain
\State $t \gets 15$, the iteration threshold
\State Initialize $A^i$, the current agent.
\State Initialize Opponent $A^{0}$, the standard DQN  agent
\While{True}
    \State \emph{interact}(\emph{Agent}: $A^i$, \emph{Opponent}: $A^{i-1}$, \emph{Env})
    \State \emph{update}(\emph{Agent}: $A^i$)
    \If {$ S\%C == 0$}
        \State average\_score = evaluate($A^i$, $A^{i-1}$)
        \If{average\_score $>$ t}
            \State \emph{Opponent} $\gets A^i$
            \State $i += 1$
        \EndIf
    \EndIf
    \State $S += 1$
\EndWhile
\end{algorithmic}
\end{algorithm}
\end{minipage}

\subsection{Linear Probing}

Linear Probing is a technique used to determine if the learned representation in layers of a neural network contains information relevant for solving a sample task. This entails using simple machine-learning models like linear regressors to map layer activations to target values~\citep{BERT}. Designing the heuristic task is entirely problem-dependent; the task often includes targets that a human expects is important toward achieving the end goal. 

In the setting of Pong, we assume that a good agent should have the ability to predict, with high accuracy, where the ball is going to land on its side of the game board. Therefore, we attempted to linearly map the agent's final layer activations (a $512$-length vector) to the landing $y$ position of the ball. A ball landing at the top of the board on the agent's side has a landing value of $0$, and a ball landing on the bottom has a landing value of $82$, the height of board in pixels. We collect a training and test set of $10,000$ frames each; frames in this set are taken $\leq 30$ steps before the ball collides with the agent's side of board. We report the MSE and $R^2$ statistic using $A^{7}$ from Chainer, Pool trained with $25$ million steps, and the standard DQN agent trained for $45$ million steps.

\section{Experiments}

In this section, we show empirical results for Chainer and Pool. We present learning curves, a new robustness metric, and the linear probing results. 

\subsection{Chainer and Pool Results}
\begin{figure}%
    \centering
    \subfloat[\centering Chainer ]{{\includegraphics[width=8cm]{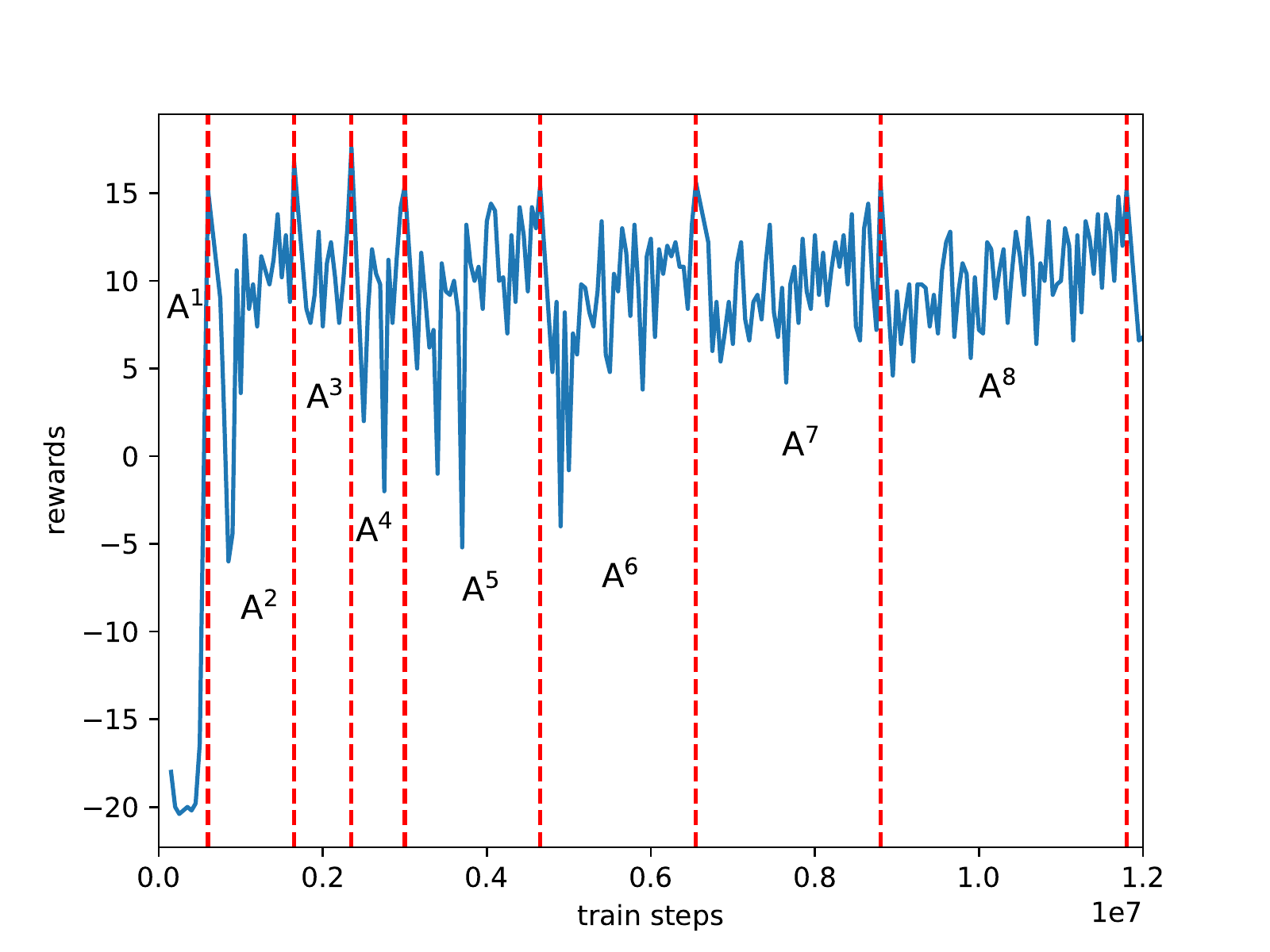} }}%
    \qquad
    \subfloat[\centering Pooling]{{\includegraphics[width=8cm]{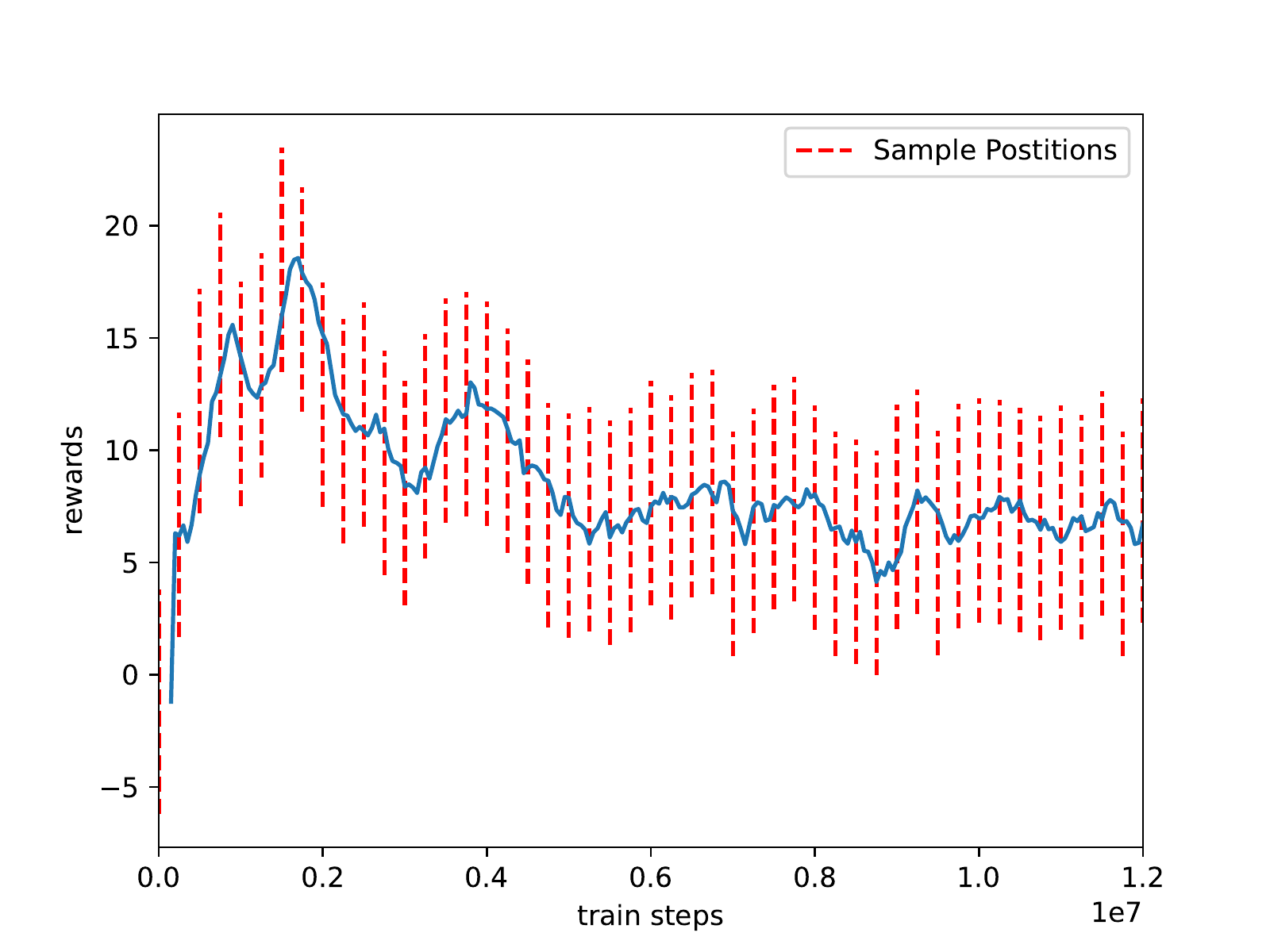} }}%
    \caption{Learning Curves for Chainer and Pool}%
    \label{fig:chainer_pool_learning_curve}%
\end{figure}

The Chainer learning curve shown in Figure~\ref{fig:chainer_pool_learning_curve}~(a) shows the game reward achieved throughout training for different iterations of the opponent. Each vertical red dashed line represents a changing of the opponent (for example, the first dashed line represents the point in time when the opponent becomes $A^{2}$, the frozen agent at the second iteration of the chain). We move on to the next agent in the chain after the current agent has converged, which we define as reaching an average score of $15$ over $10$ matches against the opponent ($21$-point game). Notice how, as training progresses, the number of steps it takes for the agent to converge against its adversary increases---indicating the strength of opponents is increasing in the chain. 

The Pooling learning curve in Figure~\ref{fig:chainer_pool_learning_curve}~(b) shows the game reward achieved throughout training for different samplings of opponents from the queue. In this graph, the vertical dashed lines represent points in time where we sample a new opponent from the pool (every $250,000$ steps). Notice that, as time goes on, the overall reward curve dips downward. That is because the agents in the pool are getting better, making it more difficult for the current agent to perform well. Also note, in comparison to Figure~\ref{fig:chainer_pool_learning_curve}~(a), this curve is not training agents to convergence. As a result, the agent finds it more difficult to adapt to the changing opponent. We show in Table~\ref{table:standard_DQN_match} how, even though the agent is challenged by these new, more powerful opponents, it retains the ability to defeat the standard DQN agent. 

\subsection{Robustness}

A truly robust agent should be hard to defeat by an opponent policy because its play is strong across the board. Thus, we define the \emph{robustness} of an agent as the difficulty for an opponent to defeat it through learning. More robust agents will have flatter adversary score curves, indicating less reward achieved by the opponent over the course of learning. We picked the scores achieved by the opponent after $6e6$ steps, after which point the learning curve was stable with no large fluctuations. 

% MLL zzz: I agree with George. If it's a metric, there should be a specific value, not just a description of a plot. Could be adversary score after 1 million steps of training.
% zzz MLL: NOTE: The labels in the graphs in Figure 2 are too small. Please make them (the labels, not the graphs) bigger. The colors (mainly the yellow) are pretty terrible, too. Too bad you weren't able to match the training steps in the two graphs, which would keep the colors down and convey a bit more information about training time vs. robustness for the two approaches.
% zzz MLL: Also, please reorder the labels in the legend so they match the order of the lines: standard DQN should go first in both plots.
% zzz: Do you think renaming 500k to 0.5M would make comparison easier?

In our robustness experiments, we tested $3$ Pooling agents ($A_{\mathrm{0.5M}}, A_{\mathrm{4M}}, A_{\mathrm{25M}}$) and $3$ Chainer agents ($A^{1}, A^{4}, A^{7}$). Later agents are consistently more robust than earlier ones, as shown by their lower lines in the graphs.
% progresses, the agents become more robust. 
In comparison, the standard DQN agent is easily defeated by an opponent, both standard DQN agent curves reach $21$ within about $1$ million steps. 
% zzz MLL: Therefore, its robustness is...?

\begin{figure}%
    \centering
    \subfloat[\centering Chainer Robustness Score]{{\includegraphics[width=8cm]{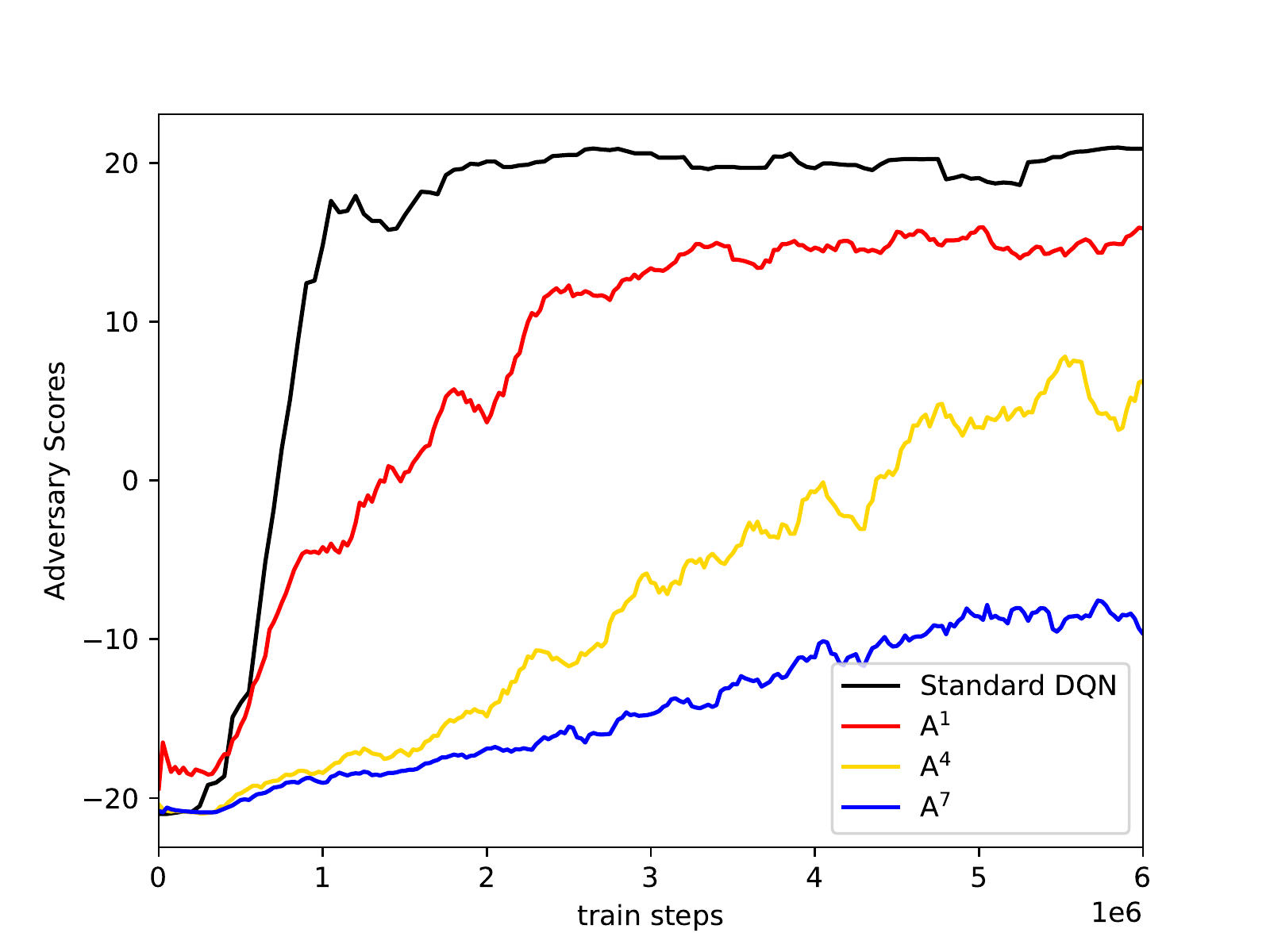} }}%
    \qquad
    \subfloat[\centering Pooling Robustness]{{\includegraphics[width=8cm]{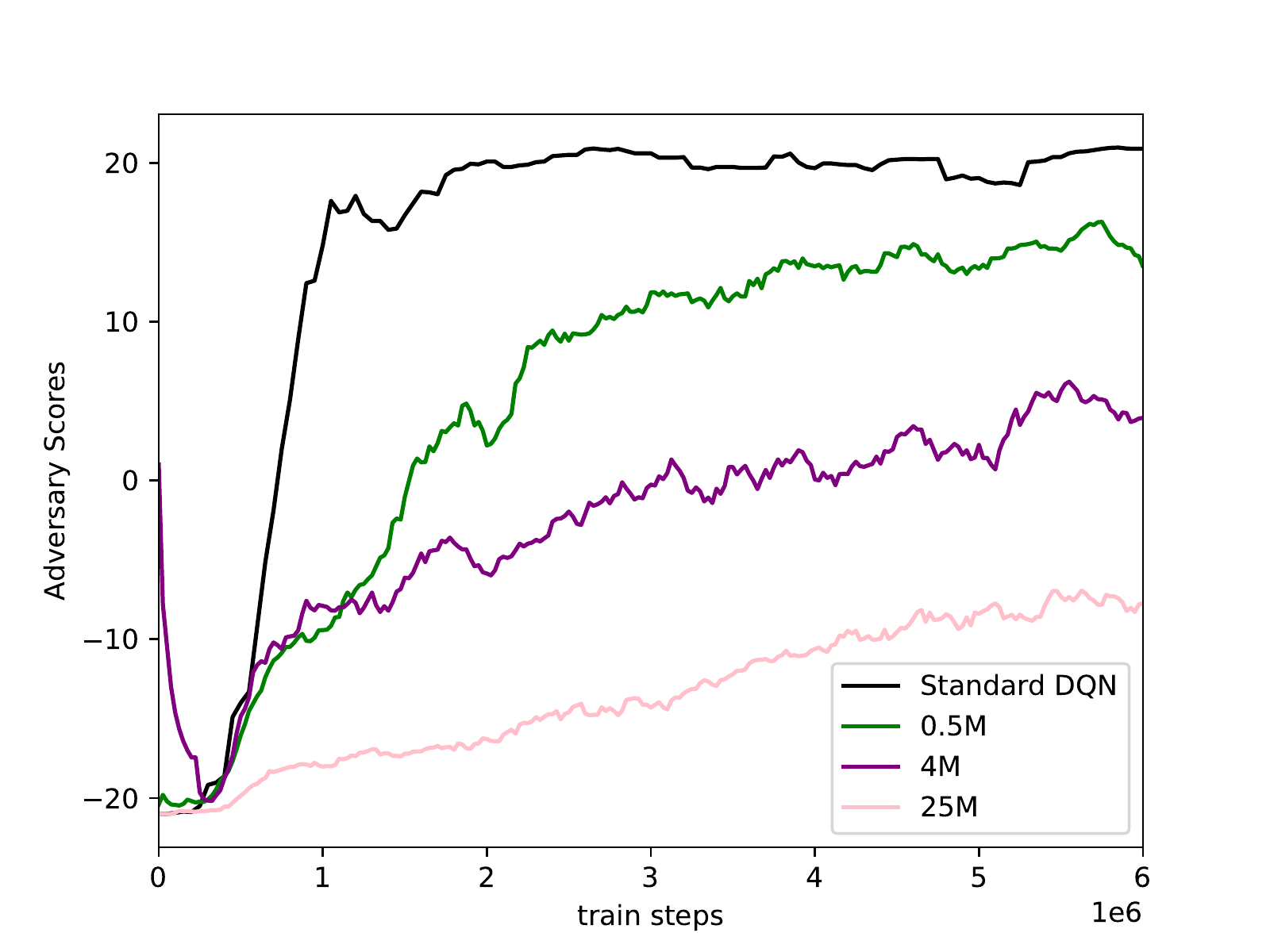} }}%
    \caption{Robustness Tests (lower is more robust)}%
    \label{fig:example}%
\end{figure}

\subsection{Linear Probing}

\begin{wraptable}{r}{6cm}
% \begin{center}
    % \begin{table}
        % \centering

        \begin{tabular}{|c|c|c|}
            Training Scheme & MSE & $R^2$ \\ 
            Standard DQN & $276.97$ & $0.41$ \\  
            Pool & $211.84$ & $0.55$\\
            Chainer & $202.41$ & $0.57$
        \end{tabular}
        \caption{Linear Probing Results}
        \label{table:linear_probing_results}

    % \end{table}
% \end{center}
\end{wraptable}

The results for Linear Probing are shown in Table $\ref{table:linear_probing_results}$. Probes for Chainer and Pool have significantly lower MSE and higher $R^2$ statistics than the probe for standard DQN agent. A regression model that randomly guesses the landing $y$ position of the ball would have MSE $\approx (82 - 41)^2 \approx 1600$. It is important to note that while the true model regressing network activations onto the landing $y$ position of the ball is likely non-linear, even using the crudest form of estimation with a linear model shows that the activations from adversarially trained agents have greater predictive power on this task, suggesting these agents more thoroughly capture the dynamics of the game.

\subsection{Adversarial Agents vs.\ Standard DQN}

An obvious question that the learning curves in Figure~\ref{fig:chainer_pool_learning_curve} raises is whether the adversarial agents can consistently defeat the standard DQN. For example, do Chainer agents in link $A^{n+1}$ remember how to defeat the standard DQN agent ($A^{0}$, trained for $45$ million steps) even after being trained to defeat more difficult opponents? 
% Can Pooling defeat standard DQN even after being trained on more difficult opponents? 
To answer this type of question, we play Pooling agents ($A_{\mathrm{0.5M}}, A_{\mathrm{4M}}, A_{\mathrm{25M}}$) and Chainer agents ($A^{1}, A^{4}, A^{7}$) against the standard DQN agent and report the 10-match average score achieved. The general trend is clear, as agents get further along in the adversarial training, they are able to still easily defeat the standard DQN agent (indicating that these gameplay skills are not forgotten). In fact, their relative strength appears to be steadily increasing. We reiterate that it is not simply a matter of the adversarial agents being trained longer---the longest training agents have just over half the experience used to train the standard DQN agent. The difference is that the adversarial agents are taught to handle a wider variety of situations, resulting in stronger overall play.
% zzz MLL: Is it possible to mention performance against the built-in Atari agent? The issue here is not that standard DQN isn't good at what it was trained to do (beat the built-in agent), and we should emphasize that if we can.

\begin{center}
    \begin{table}[h!]
        \centering
        \begin{tabular}{|c|c|c|}
            Chainer Agents & Scores Achieved\\& Against Standard DQN \\ 
            $A^1$ & $16.7$\\  
            $A^4$ & $19.5$\\
            $A^7$ & $20$\\
        \end{tabular}
        \quad
        \begin{tabular}{|c|c|c|}
            Pooling Agents& Scores Achieved\\& Against Standard DQN\\ 
            $0.5M$ & $18.2$\\  
            $4M$ & $15.1$\\
            $10M$ & $20.2$\\
            $25M$ & $20.4$\\
        \end{tabular}
        \caption{Adversarial Agents vs.\ Standard DQN Agent (trained for $45$ million steps)}
        \label{table:standard_DQN_match}
    \end{table}
\end{center}

\section{Conclusion}
These results validate adversarial training schemes as a way to produce robust agents capable of opponent generalization in Pong. Agents trained using either Chainer or Pool are capable of handily defeating a standard DQN agent. Lastly, Chainer and Pool agents are more robust (more difficult to defeat by a learning opponent) and have better informed internal activations when compared to standard DQN agents. 

\bibliographystyle{unsrtnat}

\bibliography{references}
\end{document}